\documentclass[11pt]{article}
\usepackage{style/emnlp2021}
\usepackage{times}
\usepackage{inconsolata}
\usepackage{amssymb}
\usepackage{pifont}

\usepackage{subfloat}
\usepackage{graphicx}
\usepackage{caption}
\usepackage{subcaption}

\usepackage{url}  
\usepackage{bibunits}
\usepackage{floatrow}
\usepackage{fnpct}
\usepackage{microtype}
\usepackage{soul}
\usepackage{colortbl}

\definecolor{Mycolor1}{HTML}{CCCCCC}

\definecolor{Mycolor2}{HTML}{9999FF}
\definecolor{Mycolor3}{HTML}{FF9900}
\definecolor{Mycolor4}{HTML}{FF6666}

\newif\ifcomment
\commenttrue

\newif\ifdoublespaceme
\doublespacemefalse

\usepackage{framed}
\usepackage{soul}
\usepackage{mdwlist}
\usepackage{latexsym}
\usepackage{nicefrac}
\usepackage{booktabs}
\usepackage{amsfonts}
\usepackage{bold-extra}
\usepackage{amsmath}
\usepackage{fnpct}
\usepackage{dsfont}
\usepackage{amssymb}
\usepackage{bm}
\usepackage{graphicx}
\usepackage{mathtools}
\usepackage{microtype}
\usepackage{multirow}
\usepackage{multicol}
\usepackage{xspace} %%LL-0425: added for spacing in macro text
\usepackage{comment}
\usepackage{xcolor}
\usepackage[nomessages]{fp}
\usepackage{tikz}
\usepackage{calc}
\usepackage{pgfplots, pgfplotstable}
\usepackage{filecontents}

\pgfplotstableset{col sep=comma}

\newcommand{\gem}[1]{\mbox{\textsc{gem}}}
\newcommand{\abr}[1]{\textsc{#1}}

\newcommand{\g}{\, \mid \,}

\newcommand{\hidetext}[1]{}
\newcommand{\ignore}[1]{}

\ifcomment
\newcommand{\todo}[1]{\textcolor{red}{{\bf TODO: #1}}}
\else
\newcommand{\todo}[1]{}
\fi

\ifcomment
\newcommand{\pinaforecomment}[3]{\colorbox{#1}{\parbox{.8\linewidth}{#2: #3}}}
\else
\newcommand{\pinaforecomment}[3]{}
\fi

\newcommand{\smallurl}[1]{ \begin{tiny}\url{#1}\end{tiny}}

\definecolor{lightblue}{HTML}{3cc7ea}
\definecolor{CUgold}{HTML}{CFB87C}
\definecolor{grey}{rgb}{0.95,0.95,0.95}
\definecolor{ceil}{rgb}{0.57, 0.63, 0.81}

\newlength\maxlen
\newlength\unitlen

\definecolor{exampleblue}{RGB}{189, 215, 238}
\definecolor{exampletextblue}{RGB}{47, 85, 151}

\makeatletter
\renewcommand\sectionautorefname{\S\@gobble}  
\newcommand{\figfile}[1]{2021_emnlp_weak_dpr/figures/#1}

\newcommand{\hotpot}{\textsc{HotPotQA}\xspace} 
\newcommand{\bert}{{\abr{bert}}\xspace} 
\newcommand{\nq}{{\abr{NaturalQuestions}}\xspace} 
\newcommand{\name}{\textsc{DistDR}\xspace} 
\newcommand{\beamdr}{\textsc{BeamDR}\xspace} 
\newcommand{\dpr}{\textsc{dpr}\xspace} 

\definecolor{midnightgreen}{rgb}{0.0, 0.29, 0.33}
\definecolor{darkpink}{rgb}{0.91, 0.33, 0.5}

\newcommand{\grr}{\textsc{\abr{grr}}\xspace} 

\newcommand*\colourcheck[1]{
  \expandafter\newcommand\csname #1check\endcsname{\textcolor{#1}{\ding{52}}}
}
\colourcheck{green}

\newcommand*\colourcross[1]{
  \expandafter\newcommand\csname #1cross\endcsname{\textcolor{#1}{\ding{55}}}
}
\colourcross{red}

\title{Distantly-Supervised Evidence Retrieval Enables Question Answering without Evidence Annotation}

\author{
  Chen Zhao\\
  \abr{cs}, \abr{umiacs}\\
  University of Maryland\\
  {\tt chenz@cs.umd.edu} \\\And
  Chenyan Xiong\\
  Microsoft Research\\
  {\tt chenyan.xiong@microsoft.com} \\\AND
  Jordan Boyd-Graber\\
  \abr{cs}, \abr{umiacs}, iSchool, \abr{lsc}\\
  University of Maryland\\
  {\tt jbg@umiacs.umd.edu} \\\And 
  Hal Daum{\'e} III\\
  \abr{cs}, \abr{umiacs}, \abr{lsc}\\
  University of Maryland\\
  Microsoft Research \\
  {\tt me@hal3.name} \\\newline}

\date{}

\begin{document}
\maketitle

\ifdoublespaceme
  \doublespacing
\fi

\begin{abstract}
  % !TEX root = ../../2021_emnlp_weak_dpr.tex

%Large-scale datasets have enabled significant progress on neural models 
%to open-domain question answering.
Open-domain question answering 
%finding
%evidence from a large corpora, and producing an answer
%based on it.
%
answers a question based on evidence retrieved from a large corpus.
% complex questions 
%that often require combining multiple evidence pieces to form a reasoning chain. 
State-of-the-art neural approaches require 
intermediate evidence annotations for training.
%
%make crucial assumptions that gold evidence pieces
%are available in training.
However, such intermediate annotations are expensive, 
and methods that rely on them cannot transfer to the more common setting,
where only question--answer pairs are available.
This paper investigates whether 
models can learn to find evidence from 
a large corpus, with only distant supervision from answer labels
for model training, thereby generating no additional annotation cost.
We introduce a novel approach (\name{}) that iteratively improves 
over a weak retriever by alternately 
finding evidence from the up-to-date model 
and encouraging the model to learn the most likely evidence.
Without using any evidence labels, \name{} is on par with
fully-supervised state-of-the-art methods 
on both multi-hop and single-hop \abr{qa} benchmarks. 
Our analysis confirms that \name{} finds more accurate evidence over iterations, 
which leads to model improvements. The code is available at 
\url{https://github.com/henryzhao5852/DistDR}.

% iteratively closes the distance between 
%the question and evidence in the dense space. 

\end{abstract}

\section{Introduction}
\label{sec:intro}

%\todo{a better title name and model name}
%\czcomment{Iteratively Dense Reasoning Chain Retrieval for Weakly-supervised Question Answering}

%\halcomment{would everyone agree that ODQA is exclusively about
%  factoid questions?}

Open-domain question answering (\textsc{odqa}) 
takes a question, retrieves evidence from a large corpus, and finds an answer
based on that evidence~\cite{voorhees1999trec}.  
With the help of large scale datasets, state-of-the-art approaches to \textsc{qa}~\cite[\emph{inter
alia}]{karpukhin-etal-2020-dense,zhao2021multi}
can answer both simple questions that require only a single evidence
piece (i.e., one passage);
and more challenging multi-hop questions: computers must
 jump or ``hop'' from passage to passage (we call these passages {\bf evidence pieces}) , building a
reasoning chain to find the answer.
%
%We call these clues {\bf evidence pieces}, and computers must put
%these puzzle pieces together to answer these complex questions.

\begin{figure}[!t]
  \centering
  \includegraphics[width=0.99\linewidth]{\figfile{ex_fig.pdf}}
  \caption{A multi-hop question example from \hotpot{} that
    requires finding multiple evidence pieces to form a reasoning chain (\underline{Sang-Wook Cheong} $\rightarrow$ \underline{Rutgers University}). 
    \textcolor{red}{Red}: Text that overlaps between question and evidence piece;
    \textcolor{blue}{Blue}: Span that matches the answer.
    State-of-the-art systems use evidence labels 
    for training, but acquiring labeled evidence pieces is expensive. 
  }
  %\jbgcomment{Shouldn't this be ``middle name'', otherwise it
  %  doesn't make sense.}

  %\jbgcomment{This also doesn't seem like a great example because
  %  Sela Ward's page does include the string Bobbi Bacha, so it
  %  would be possible with a single hop.}
  \label{fig:ex}
\end{figure}

% that require chaining multiple evidence pieces.
%\halcomment{i feel like there should be some cite for open domain QA that's more than 4 years old -- like 90s or 2000s. maybe jimmy lin? or a book?}
%%czc516 edited, it's trec report
%to form a reasoning chain. 
%
%answering
%a question given a large textual corpora. With the help of large 
%scale datasets, state-of-the-art approaches 
%can not only answer simple questions that need single evidence piece,
%but also complex questions that
%require combining multiple clues to form a reasoning chain. 
%that ``hop'' between text 
%pieces to build a reasoning chain.
%, or 
%\emph{multi-hop reasoning.}

%This task is often called 

%\halcomment{i might drop the retreiver-reader until later, it's not really relevant here in setting up the problem}
%\czcomment{edited}
%
%tackling both single-hop~\cite{kwiatkowski2019natural} and multi-hop
%questions~\cite{yang+18b}.

%\jbgcomment{say a few words to explain connection between entities}
%% cz0513 edited 

State-of-the-art (\abr{sota}) methods,
%%cz0909 remove the following since we target both single hop and multi hop.
%for such complicated \abr{qa} tasks, 
however, are trained with \emph{all}
of the intermediate evidence pieces (e.g., in
Figure~\ref{fig:ex}, the evidence pieces for 
\underline{Sang-Wook Cheong}'s workplace which point you to
\underline{Rutgers University}'s location) needed for the answer.
%\halcomment{i'm not sure eveyone will realize what breadcrumbs is referring to. i'd just be explicit about what is meant.}
%\czcomment{edited}
%
Creating such intricate training data is expensive. For example, 
\citet{kwiatkowski2019natural} use additional experts to justify the correctness 
of annotated evidence.
The annotation protocol is even
more nuanced for multi-hop questions. For example, 
%to ensure multiple evidence pieces are used, 
\citet{yang+18b} ask annotators 
to write multi-hop questions based on two linked Wikipedia passages 
as a pre-defined reasoning chain, which 
creates dataset artifacts~\cite{min2019compositional}.
%\halcomment{i'd be explicit about what makes them tricky? is it that it's ill-defined? 
%that there are often multiple good middle hops? somethinge else?}
%\czcomment{edited}
%
While plenty of 
%plain 
question-answer 
pairs are available without evidence labels,
%(and are often available organically), we cannot directly train \abr{sota}
we cannot directly train \abr{sota} models on such data. 
%\halcomment{i don't understand this last sentence}
%\czcomment{Now i also find it ambigious, let's talk during meetings.}

%\halcomment{i'd make the figure more explicit: P1/P2/P3 -> Evidence Piece 1, ... 2, ...3; also why is the whole "Sela Ann Ward" highlighted in the last piece and not just Ann? also explain the color coding here :)}

%\czcomment{I add the first half of the following paragraph, as Marine mentions 
%in paper clinic that it's hard to understand what our setting is and why we propose this method. }

%\jbgcomment{The first time you define ``evidence piece'', define it clearly}

Our work focuses on training \abr{odqa} systems without these expensive
annotations (Section~\ref{sec:setting}): we only start with a question-answer
pair.
With that starting point, we use distant supervision to infer which
evidence helps us get to the answer.
The technical challenge is how to find these evidence from millions of
candidates.
Previous methods~\cite{joshi-etal-2017-triviaqa, cheng-etal-2020-probabilistic}
use term matching (e.g., \abr{tf-idf}) for evidence retrieval, 
but their goal is a single piece of evidence: linking a question to a
passage.
As shown in Figure~\ref{fig:ex}, the key to finding some evidence
pieces does not appear in the question: for example, you only know to figure
out that \underline{Rutgers University} is in New Jersey after
learning where Professor Cheong works.
Fortunately, navigating to an answer given a question 
%and preliminary
%results 
from a search engine is not impossible: humans do it every
day, building on their existing knowledge toward the
answer~\cite{russell-19}.
With each round of searching to find additional clues, human users
accumulate information to find the
right answer.
This paper creates a computational approach (\name{}) that 
can use similar techniques to find
the evidence needed to answer a given question.

%Our approach, \name{}, builds on recent advances in 
%dense retrieval~\cite{karpukhin-etal-2020-dense}, which creates custom
%vectors for a query and evidence.
%
%This framework allows us to encode the pieces of information we
%collect from each of the hops into query vectors.
%
%While recent work shows that this can also aid both single hop questions~\cite{xiong2020approximate}
%and multi-hop questions~\cite{xiong2020answering}, these approaches
%still assume the necessary hops are known in advance.

%\jbgcomment{Might be too much detail on model below}

%\czcomment{edited, guess partically addressed}

%\halcomment{right now if i don't know those papers already, i can't make much of this paragraph. 
%i would maybe swap the order of the previous and following paragraphs, 
%so you first give the high level idea of what's new, then say the context in which you're applying it.}
%Our approach, \name{},  
\name{}
starts with a weak retriever then 
iteratively improves it by 
finding more useful evidence (Section~\ref{sec:model}).
%
%We instead iteratively improve a weak evidence-finder 
%
%
%using Expectation–maximization~(\textsc{em}) approach. 
%Through iterative process, our method increasingly retrieves 
%more reliable evidence from the corpus, 
%which is used to improve the evidence-finder (Section~\ref{sec:model}). 
%
%\halcomment{i'm not sure everyone will know what a reader is}
%%
%
%\halcomment{weak supervision i think isn't defined above, define it here}
%
%
Specifically, we model evidence as a latent variable and 
develop a hard-\abr{em} algorithm that alternates between using the up-to-date retriever to find  
evidence (hard E-step) and 
updating the model parameters to further encourage the most useful evidence
in the next iteration (M-step).
%assuming the retrieved evidence is a
%true solution (M-step). 
%\jbgcomment{I think this needs to be a little
%stronger to show how things actually improve.  Something like: ``and
%updating the retriever's parameters to favor the most useful evidence
%in the next interation''}

To implement this idea, we need a trainable retrieval system.
We use dense retrieval~\cite{lee-etal-2019-latent} as our retriever, which uses a
neural network to encode the evidence pieces we
collect at each round into query vectors.
\name{} provides iterative feedback within the context of a \abr{qa}
system to guide the encoder to better find evidence pieces.

%\halcomment{it's not totally clear to me from this description what is being treated as the latent variable in EM.
% i think it might be worth being explicit what are the parameters being optimized, and what latent variable is marginalized over.}
%\czcomment{I added one sentence}

%\czcomment{The claim of experiments could make it stronger..}

%\todo{ Add add a few more sentences about our analysis.}
We evaluate \name{} on \abr{odqa} benchmarks, 
including  
both single-hop questions~\cite[\nq{}]{kwiatkowski2019natural}, where the evidence is the target passage,
and multi-hop questions~\cite[\hotpot{}]{yang+18b}, where the evidence is a chain of passages.
\emph{Without} using 
any annotated evidence labels, \name{}'s accuracy, according to several measures, is \emph{on par}
with fully-supervised state-of-the-art approaches on both benchmarks (Section~\ref{sec:experiment}).

%\jbgcomment{does it take too much space to list the measures?}

%on both
%\hotpot{}~\cite{yang+18b}, the multi-hop \abr{qa} benchmark,
%and \nq{}~\cite{kwiatkowski2019natural}, the single-hop \abr{qa}
%benchmark.

%\emph{on par} results to fully-supervised approaches.
%On multi-hop \abr{qa} benchmark, \hotpot{}, without using 
%any annotated evidence labels, 
%\name{} matches \beamdr{} and \abr{mdr},
%state-of-the-art dense retrieval methods under 
%full supervision. On single-hop \abr{qa} benchmark, \nq{}, 
%\name{} is competitive to dense retrieval 
%approaches using annotated passages as labels,
%while outperforms the weakly-supervised conuterparts
%(Section~\ref{sec:experiment}). 

%\jbgcomment{Analysis of what?  This is too vague.}

%To better understand \name{}, we conduct analyses and ablation studies. 
Our analyses confirm the intuition that over iterations, \name{} selects more accurate 
evidence, which in turn improves the
model.
Although some retrieved evidence from \name{} does not match the annotation, 
it gives useful training signal, as \name{} finds
\emph{alternative} evidence---another advantage
of an automated approach (Section~\ref{sec:analysis}).
For example, you can connect \underline{Sang-Wook Cheong} to New Jersey
 through another Wikipedia page 
(\underline{History of Rutgers University} in Figure~\ref{fig:ex}).

%The rest of paper is organized as follows: we introduce weakly supervised \abr{qa}
%task in Section~\ref{sec:setting}, present our model in Section~\ref{sec:model}, 

 %\jbgcomment{Include section forward points}

 %\czcomment{I put section points  inline }

\section{Why Weakly-Supervised ODQA}
\label{sec:setting}

Our task is to answer questions over large textual corpora. 
Our approach is generally applicable to both single-hop 
and multi-hop questions.
%We
%motivate from answering complex questions (e.g., multi-hop) that require
%multiple evidence pieces (each evidence piece is a passage) to find the answer, but our apporach is also 
%applicable to simple questions where single evidence piece is relevant for answering the question.
We use Wikipedia as the knowledge source, but we do not use the metadata such as hyperlinks,
to ensure our method can apply to any corpus (e.g., ClueWeb). 

State-of-the-art approaches on \abr{odqa} 
%usually include two components, 
%a retrieval system that outputs a small amount of text, then a reading comprehension
%system that extracts an span as the output answer. However they 
mainly focus on the \emph{fully-supervised}
setting, where the question, answer, and evidence are given in training. Figure~\ref{fig:ex}
shows an example multi-hop question with answer and annotated evidence.  
%
%open-domain QA is widely studied in academia and 
%is used on commercial systems such as search engines. 
%However there's a discrepancy
%between academic benchmarks and real world data: benchmarks are often annotated with 
%single gold evidence, which provides 
%full supervision in model training, while real
%data only includes the question answer pair. 
%\halcomment{it's not clear to me what "real data" here means? 
%do you mean data that commercial search engines use? i'd bet they annnotate the heck out of their data :).}
%
The fully-supervised setting simplifies model training but has three 
major challenges:
(1) Annotation is costly: the question--answer pair is widely available
in the real world, but
the evidence (\underline{Sang-Wook Cheong} $\rightarrow$ \underline{Rutgers University} in Figure~\ref{fig:ex}) 
requires human annotation, which is expensive to get, especially for complex questions; 
(2) Domain generalization: the labeled evidence is only on a single corpus, generalization
to other corpora (e.g., moving from Wikipedia to medical records) is non-trivial;
(3) Alternative evidence: there are often multiple correct evidence candidates 
in the corpora (e.g., in Figure~\ref{fig:ex}, both \underline{Sang-Wook Cheong} $\rightarrow$ \underline{Rutgers University}
and \underline{Sang-Wook Cheong} $\rightarrow$ \underline{History of Rutgers University} are correct), but only one of them is annotated as ``gold evidence''.
Therefore, we explore the more common but challenging \emph{weakly-supervised} setting
which only requires question--answer input pairs. 

Formally, given a question~$q$ and a textual corpus with~$D$ passages, 
our task is to first find a small subset of relevant passages 
%(with size $k$ evalauted on top-$k$ retrieval accuracy) 
as evidence~$z$ to the answer, where each evidence combines evidence pieces~$z=z_{1}, ..., z_{n}$ 
with length~$n$ ($n=1$ for single-hop questions),
then find a span $a$ from evidence $z$ as an answer. 
We focus on the weakly-supervised setting,
where training examples consist only 
of question--answer pairs $(q, a)$,
we gather evidence $\hat z$ from corpus as the distant 
supervision signal, based on the assumption that 
the presence of the answer $a$ (source of distant supervision) in an
evidence piece
implies that the evidence is needed to answer the question.
This is contrast to the fully-supervised setting, where the gold
evidence $z^{*}$ is also given during training.

%\czcomment{I add a sentence for the definition of distant supervision, not sure if that's clear though.}

%\halcomment{you should probably also formally define the fully supervised setting, since you'll compare with it}

%\czcomment{edited}
\section{Weakly-Supervised ODQA with \name{}}
\label{sec:model}

%\czcomment{not sure if we need an additional algorithm figure showing each steps}

\begin{figure*}[!t]
  \centering
  \small
  \includegraphics[width=0.99\linewidth]{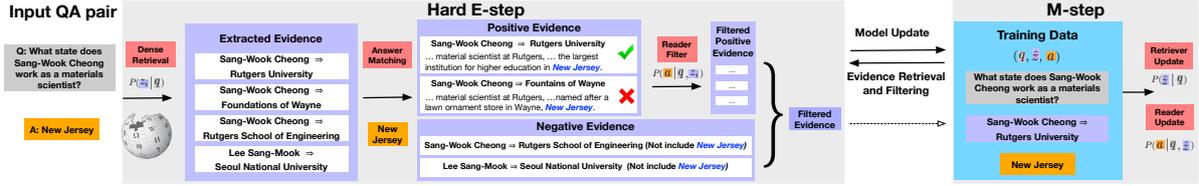}
  \caption{Our \name{} model, with question: \colorbox{Mycolor1}{q}, evidence: \colorbox{Mycolor2}{z}, answer: \colorbox{Mycolor3}{a}, model component: \colorbox{Mycolor4}{Dense Retrieval}. 
  \greencheck indicates reader output matches the correct answer, thus the positive evidence is kept (otherwise filtered out, presented by \redcross). 
  Left: at E-step, \name{} finds the most relevant evidence using the current dense retriever on the training examples, 
   uses both answer string matching and 
  reader filter to form positive and negative evidence. Right: at M-step, \name{} updates 
  both retriever and reader components using the training data from
  E-step as distant supervision.}
%\jbgcomment{Is there a way we can get that we're looking for New
%  Jersey in the positive evidence and throwing out some examples in
%  the filtering step?  Even if it makes the figure bigger, I think
%  that would help.}
  \label{fig:model}
  \end{figure*}
%We follow 
%the retriever-reader framework \halcomment{define this framework}, while our focus is on the reteivel component, which
%aims to find a small subset of relevant text (with length $k$ evalauted on top-k retrieval accuracy) 
%as evidence to find the answer.
%\czcomment{there are still mismatch between distant supervised and weakly supervised setting, need to make it consistent}
%\jbgcomment{Why is evidence called both $e$ and $z$?  Call them different names if they are different.}

We present \name{}, a unified framework for weakly-supervised
\abr{odqa}.
\name{} is trained by retrieving evidence from a large corpus with
distant supervision.
%
%In the fully-supervised setting, the annotated evidence~$z$ is given, 
%the goal is to learn the model parameter 
%$\theta$ that takes an input question~$q$,
%outputs the evidence $z$, and the reader further takes the 
%question $q$ and evidence $z$, output an answer span.
%

\name{} follows the retriever--reader framework for \abr{odqa}, using 
dense retrieval to find evidence (Section~\ref{subsec:dr}).
%which we review in Section~\ref{subsec:dr}. 
%Through matching passages 
%and questions with dense representations, 
%\abr{dr} removes the burden of little textual 
%overlap between question and passage 
%for unsupervised term matching methods. 
However, 
%unlike previous work that directly uses evidence label $z^{*}$ to train models, 
in our approach, we only have questions~$q$ and answers~$a$,
so we need to induce evidence~$\hat z$ for training our retriever.
%
%
%Since the search space is large, i
We fully expect that our initial retriever will struggle to find evidence.
However, if it can find \emph{some} useful evidence, we can encourage it to
follow the same clues to more evidence that can answer questions.
This intuition is the foundation for our iterative approach (Section~\ref{subsec:iterdr}) for evidence retrieval: an initial retriever
attempts to find evidence (Dense Retrieval).
If the evidence contains the answer (Answer Matching), 
we label it as positive evidence (otherwise it's negative evidence);
then we use the retrieved evidence as labels to retrain the retriever and reader (Model Update). 
While this idea forms the basis of our algorithm, it can be led astray
by false-positives: evidence that contains the answer but is
irrelevant to question.
Returning to our running example, while ``named after a lawn ornament store in Wayne,
\underline{New Jersey}'' has the state where Professor Cheong works,
it is irrelevant to condensed matter physics.
Thus, we use a reader to filter spurious evidence (Reader Filter) and
keep \name{} on target.

\subsection{Preliminary: Fully-Supervised ODQA}
\label{subsec:dr}

This section reviews state-of-the-art systems for fully-supervised \abr{odqa},
where a dense retriever finds evidence from a large corpus, 
and a reader---multi-tasked with evidence reranking and span extraction---outputs a span as the answer.

%\jbgcomment{Reformat math, don't write words like $this$.  Operators should be in text blocks.}

%\jbgcomment{$q$ is being used for both the original question and then
%  the encoding $q$ for MIPS.  This is confusing.}

\paragraph{Dense Retrieval} Dense retrieval~\cite{lee-etal-2019-latent}
is based on a dual-encoder architecture, which uses 
\abr{bert} to represent both the query~$q$ and the passage~$p$ with dense vectors.
% $\mbox{Enc}_{Q}(q)$ and $\mbox{Enc}_{P} (p)$.
The model learns a scoring function (e.g., dot product) between
question and passage vectors:
\begin{equation}
    f(q, p) = \text{sim}(\mbox{Enc}_{Q}(q), \mbox{Enc}_{P}(p)).
    \label{eqn:sim}
  \end{equation}
These models are highly scalable, since passages can be encoded
offline, and 
are efficiently retrieved over maximum
inner product search (\abr{mips}) with the query~\cite{shrivastava2014asymmetric}.
%\halcomment{typographic: i put periods inside paragraph\{\}s \emph{or} i include the text as part of a sentence, but i don't want to clash jordan's stylistic preferences}

%\halcomment{typographic:  there are a lot of Es floating around. E for embedding, e for exponential, and presumably E for expectation (see comment below). would it be too weird to not use E for embedding? its almost never used (except here), so i would be tempted to go with $\text{emb}_Q(q)$ or something isntead. i also would use $\exp[...]$ rather than $e^{...}$; I think it looks a lot nicer when there are lots of super/subscripts going on.}

\paragraph{Multi-step Evidence Retrieval}
We apply multiple dense retrieval steps to find evidence with a 
sequence of evidence pieces (each piece is a passage)~${z_{1},\dots,z_{n}}$, 
After each retrieval step, we create the new query by concatenating retrieved evidence pieces to the original question.
Specifically, at retrieval step $t$, 
we form a new query~$q_{t}$ by appending our already 
retrieved evidence pieces to the original question
\begin{equation}
  q_t=[q;z_{1};\dots;z_{t-1}],
  \label{eq:append}
\end{equation}
and retrieve a new evidence piece $z_{t}$. 
During inference, a beam search finds the  top-$k$ evidence, 
where the score is the product of individual evidence pieces' score.
For training, given
a positive evidence $ z^{+} \equiv z^{+}_{1},\dots,z^{+}_{n}$ (In Figure~\ref{fig:ex}, $z^{+}_{1}$: \underline{Sang-Wook Cheong}, $z^{+}_{2}$: \underline{Rutgers University}) and a set of negative evidence $Z^{-}$, we use negative
log likelihood loss (\abr{nll}) over each step:
\begin{align}
  L&(q, z^{+}, Z^{-})=   \\\nonumber
  &\sum_{t}{\frac{\mbox{exp}^{f([q;z_{t-1}^{+}], z^{+}_{t})}}{\mbox{exp}^{f([q;z_{t-1}^{+}], z^{+}_{t})} + \sum\limits_{z^{-}\in Z^{-}}{\mbox{exp}^{f([q;z^{-}_{t-1}], z^{-}_{t})}}}}.
  \label{eqn:loss}
\end{align}

%\jbgcomment{The convention is to write \texttt{BERT encoding input} in typewriter font, not math.}
%\frac{e^{f([q;p_{t-1}^{+}], p^{+}_{t})}}{e^{f([q;p_{t-1}^{+}], z^{+}_{t})} + \sum_{j=1}^{m} {e^{f([q;z_{j, t-1}], z^{-}_{j, t})}}}}
\paragraph{Evidence Reranking and Span Extraction}
After the retriever, \abr{qa} systems 
need a reader, which combines 
evidence selection (reranking) and span extraction.
Unlike retrieval, 
%which represent questions 
%and evidence separatively, 
the reader encodes pairwise information between question and evidence, thus
giving a more accurate (but slower) prediction.
Following \citet{karpukhin-etal-2020-dense}, we use a \abr{bert} encoder with input formatted 
as  \texttt{[CLS] question [SEP] title$_{1}$ [SEP] evi$_{1}$ [SEP] ... title$_{n}$ [SEP] evi$_{n}$[SEP]},
where \texttt{[CLS]} and \texttt{[SEP]} are special tokens, and each $\texttt{evi}_{t}$ is an evidence piece $z_t$.
First, we use \texttt{[CLS]} token's representation $\textbf{U}_{ \texttt{[CLS]}}$ to estimate the 
probability that the collected evidence~$z$ contains the answer:
%to answer the question:
\begin{equation}
  P(z{\g}q) = \text{softmax}(\textbf{U}_{\texttt{[CLS]}}^{\intercal}\textbf{w}_{\text{rank}}),\\
  \label{eqn:rerank}
\end{equation}
where $\textbf{w}_{\text{rank}}$ is a weight vector. 
Then answers are predicted
%\jbgcomment{Is this notation distinguishing vectors, matrices, and scalars?}
%from the \bert{} output representations $U$ 
with a span start and end classifier:
\begin{align}
    P(\text{start}{\g}q, z) &= \text{softmax}(\textbf{U}\textbf{w}_{\text{start}}),\\
    P(\text{end}{\g}q, z) &= \text{softmax}(\textbf{U}\textbf{w}_{\text{end}}),
    \label{eqn:spn}
  \end{align}
where $\textbf{w}_{\text{start}}$ and $\textbf{w}_{\text{end}}$ are weights.  From the
highest-scored evidence $ \hat z$, we select the answer with the highest span probability
$P(\text{start}{\g}\hat z, q) \times P(\text{end}{\g}\hat z, q)$. 
The training objective is the log-likelihood of the positive evidence for reranking and 
maximum marginal likelihood over all spans in the positive evidence for span extraction.
%\halcomment{i don't totally get this. you select the answer how? 
%the one with the highest p(start) and p(end)? what if the highest p(end) comes before the highest p(start)?}

%\halcomment{$w_{...}$ are undefined.}

%\halcomment{is "bert's [cls] representation $U_{cls}$ the OUTPUT representation?}
%% cz: yes, edited

%\halcomment{I don't understand eq 4 and 5. i'm guessing this is going over all words in z or something and computing p(start/end) for each of these words? what is U? what is being indexed over? once you have these start/end probabilities, how do you extract a single span?}
%% cz0513: addressed

%\halcomment{more of a discussion point, i personally have always found it a bit weird when papers talk about EM and then only deep 
%in the belly of the paper do you find out that they're actually doing alternating optimization (aka hard EM) instead of full EM. 
%right now, I think this is where we find this out. 
%that makes me a bit uncomfortable, but i also understand that saying you do "EM" gets your paper accepted, so....}

\subsection{Learning \name{}: Hard-EM }
\label{subsec:iterdr}

%\jbgcomment{Below paragraph needs a topic sentence}

This section introduces training \name{} with weak supervision.
We treat the evidence~$z$
as a latent variable (Figure~\ref{fig:model}) with a multinomial distribution (parameterized by probability of 
selecting correct evidence from the retriever) 
over solution set $Z=D\times n$, where  
$D$ is the corpus size and $n$ is the number of retrieval steps. 
Formally, given the question~$q$ and answer~$a$, our goal is to maximize the probability of finding correct evidence $P(z{\g}q)$
(For single-hop questions, $z$ is retrieved with single step;
for multi-hop questions, it takes multiple steps to find $z$:
after the first retrieval step, the query~$q$ is not just the original
question but also appended evidence as in Equation~\ref{eq:append}.),
and selecting an answer from the question $P(a{\g}q, z)$.
%
%
%Since it's intractable to enumerate all evidence candidates,
We use expectation maximization (\abr{em}) to infer the latent
variable $z$.
%%cz0908: i deleted the at each step.
%
%cz0908: I deleted given the combined query and evidence, z_i is an evidence defined above (not evidence piece, which is p_i).
We compute the 
likelihood of each $z$ given~$q$, yielding a vector of estimates $\hat z$ (E-step);
then update the model parameters based on $\hat z$ (M-step).
Since it's intractable to enumerate all evidence candidates to compute the expectation, 
we adopt hard-\abr{em}~\cite{samdani2012unified} and approximate the E-step 
by picking the most likely solution as $\hat z$.
We pass over all questions in the training set and repeat this 
process for multiple iterations until the model converges.
% \jbgcomment{Be precise about
%  ``repeat this process'', e.g. pass over all queries}

%the most likely solution $\hat z$  from the retriever $p$ (E-step), 
%and updates the model parameter (M-step) to further encourage 
%the prediction. 

%\halcomment{i think a better cite for hard EM might be: Samdani, Rajhans, Ming-Wei Chang, and Dan Roth. "Unified expectation maximization." Proceedings of the 2012 Conference of the North American Chapter of the Association for Computational Linguistics: Human Language Technologies. 2012. ... there's gotta be something older but honestly i don't know the right citation. i thought radford neal had something but i couldn't find it.}

%\halcomment{i think you should write out the math. what is the p() you're tryingt o maximize, 
%how does this look under EM -- like what would be thea ctual E and M steps, and then say that you approximate E with hard-em.}
%%cz 0513 edited

%\jbgcomment{Use $\g$ macro, not pipe}
%\todo{mention that there are multiple ways, we choose the hard-EM one}

%\jbgcomment{Talk about the distribution over evidence explicitly.  What is in the object, why do you take the most likely out of it.}
%\czcomment{edited above}

\paragraph{Hard E-step} 
%In \abr{em} framework, e-step is used to computed the likelihood od 
At the hard E-step, for each question~$q$ in the training set we find 
the most likely estimated evidence~$\hat z$. This is implemented by
multiple retrieval steps for multi-hop questions, specifically at
each step $t$:
\begin{equation}
    \hat z_{t} = \arg\max_{\hat z \in Z}P(\hat z_{t}{\g}q,\hat z_{1},...,\hat z_{t-1});
    \label{eqn:estep}
  \end{equation}
and single-step retrieval for single-hop questions.
We use an up-to-date retriever to find the top-$k$ evidence from
corpus (with beam search for the $k$-highest scoring chains, Section~\ref{subsec:dr}).
%This is implemented by
%multiple retrieval steps (with beam search for $k$-highest scoring chains)
%for multi-hop questions; or 
%single-step retrieval for single-hop questions (Section~\ref{subsec:dr}).
%We implement this computation first by
%by beam search from an up-to-date
%multi-step dense retriever 
%(we form a new query after each retrieval step $t$ with 
%already retrieved evidence pieces $p_1, ..., p_{t-1}$); 
%or single step for
%single-hop questions to find top-$k$ evidence from corpus
%(Section~\ref{subsec:dr}).  
%\jbgcomment{Make this clearer, beam search
%over what with what objective}
%and also refresh the most challenging negative chains to the current model~\cite{guu2020realm,xiong2020approximate}.  
%We first conduct beam search to find top-K evidence using the
%up-to-date multi-step dense retriever described in
%Section~\ref{subsec:dr} (or single step if the question is single
%hop),

Given the retrieval output, we will eventually need to retrain
the retriever.
This requires knowing which evidence is useful and which is not.
As a proxy, we look for the \emph{answer} to split the top-$k$
candidate evidence into positive (has the answer)
 $\hat Z^{+}$ and negative $\hat Z^{-}$ evidence (lacks the answer) sets.
And as a by-product, we generate the most
challenging negative evidence at each iteration, which makes
 training more robust~\cite{guu2020realm}. 
 %\jbgcomment{What does
 %  ``update'' mean, this is unclear.  Be precise.}

%\halcomment{what is $Z$? what is $\theta$? is this stochastic EM (ie one data point at a time) or do you do max over all data points at once?}
%% cz0513: edited, i do max over all data points at a time

\paragraph{Evidence Filter} 
Although using the answer to filter evidence ensures
that the positive evidence contains the answer, 
it does not always mean this evidence 
is relevant. For example, the answer 
to ``Who played in the most world series \abr{mlb} games''
is \ul{New York Yankees}, but 
``\ul{New York Yankees} is an American professional baseball team'' is 
not the \emph{correct} evidence. This issue is more pronounced at the 
beginning of the process when the retriever is weaker.
%and the answers are common nouns are numbers, since it's easier to match the evidence.
%
To mitigate this issue, the reader filters spurious
positives: if it does not believe \ul{New York Yankees} is the answer
to the question, the evidence is not usable.
Specifically, for each evidence 
$\hat z^{+}$ in the positive evidence set,
%that contains the answer, 
the current reader model outputs the most likely
answer~$\hat a$.
% given~$(q, \hat z^{+})$.
%
We only keep positive evidence if~$\hat a$ matches the correct
answer.

\paragraph{M-step}
Now that we have our estimated evidence~$\hat z$, including both the highest scored positive evidence
$\hat z^{+}$ (after filtering), which we assume is a true solution, and $\hat Z^{-}$ as negative evidence set. 
We have~$(q, \hat z, a)$ for each training example to 
update both the retriever~$P(\hat z{\g}q)$ and reader~$P(a{\g}q, \hat
z)$ (Section~\ref{subsec:dr}).

%\halcomment{what happened to $\hat z_+$ and $\hat z_-$? here it looks like just $\hat z$ is used?}

%we update parameters of both the retriever and reader model parameters based on the 
%retrieved evidence chain $\hat z_{+}$ and $\hat z_{-}$ from the E-step. 

\section{Experiments}
\label{sec:experiment}

%\jbgcomment{Can we get more specific section and subsection titles?}

%\todo{I might need to add statistics of the dataset}

In this section, we evaluate \name{} on both multi-hop 
and single-hop \abr{qa} benchmarks.  \name{} is generally 
applicable to both questions by adopting different 
evidence retrieval steps.

%\halcomment{maybe re-emphasize why distdr is applicable in both settings.}

\subsection{Datasets}
We evaluate on two datasets (statistics in Table~\ref{tb:statistics}), \hotpot{} and \nq{}.

\paragraph{\hotpot{}}~\citep{yang+18b} is a multi-hop \abr{qa} benchmark,
 where intermediate hops have been annotated by hand. 
We focus on the full wiki setting, where the corpus is first passage
of all Wikipedia pages ($5.23$ million passages). 
We do not use its supporting facts (evidence) annotation in our setting.
%The original dataset provides annotated relevant connected Wikipedia titles as the evidence labels, 
%while we focus on the distantly supervised setting where 
%each training example only consists of the question-answer pair. 
We only use its bridge questions subset, which are designed to be multi-hop. 
\hotpot{} also includes comparison questions that compare properties of two question entities, 
but its yes/no answers are beyond the scope of this paper, as we cannot get distant supervision
from them.

%\hotpot{} includes both bridge and comparison
%questions \halcomment{will reviewers know what these mean? at any rate i don't really understand what follows}, we only use the bridge questions, as 
%finding the evidence for comparison questions only require entity linke, and comparison questions 
%include the yes/no answer, which beyonds the scope of distant supervision setting. 
%, where a bridge entity is required to find the answer, and the comparison questions,
%which compare two entities from the same categories. We only focus on the bridge questions, as 
%the comparison questions include yes/no answer, which beyonds the scope of distant supervision
%setting. 

\paragraph{\nq{}}~\citep{kwiatkowski2019natural} is a \abr{qa} benchmark,
which mainly includes single-hop questions. 
Besides questions and answers, \nq{} also annotates passages as evidence,
but we do not use it in the weakly-supervised setting. 
We follow \citet{karpukhin-etal-2020-dense} and use all of Wikipedia as 
a corpus, split into $100$-token chunks ($21$ million passages).
%\footnote{While } 
%\halcomment{i don't get this footnote}

%\todo{add more implementation details }

\subsection{Evaluation Metrics}

%\jbgcomment{Why are you calling them passage sequences and not evidence?}

On \hotpot{}, we evaluate the retrieval component on ten evidence (chains), where each
sequence has two passages.
%\halcomment{is it obvious what passage chains means at this point?}.
For retrieval, we follow \citet{zhao2021multi} and report \textbf{answer recall} 
(the fraction of questions with the answer string in the retrieved passages),
\textbf{passage recall} (if at least one gold passage is in the retrieved passages),
and \textbf{chain recall} (if both gold passages are included in the
retrieved passages) on the dev set.
For the reader, we first use the same metrics as above on the top ten chains
reranked from top-100 retrieval results for reranking. 
Then we report an exact match (\abr{em}) score on answer spans.
On \nq{}, we report answer recall on top-$k$ passages ($k=1, 20$) from
the retriever, and exact match (\abr{em}) on answer spans on test set,
following~\citet{karpukhin-etal-2020-dense}.

\subsection{Compared Methods}

On \hotpot{} retrieval, we compare \name{} with unsupervised
\abr{tf-idf}, and two recent state-of-the-art multi-step dense
retrieval methods---\beamdr{}~\cite{zhao2021multi} and
\abr{mdr}~\cite{xiong2020answering}---under full supervision.  For the
reader, we first compare \name{} with \beamdr{} and two other top
leaderboard entries, Transformer-\abr{xh}~\cite{zhaotransxh2020} and
\abr{grr}~\cite{asai2020learning}, both of which use Wikipedia
hyperlinks to find candidates.  For fair comparison, all approaches
use \bert{}-base as pre-trained model, and we use the released model
checkpoints to do inference on bridge question subsets.\footnote{\abr{mdr}
  uses RoBERTa-base for retrieval, and \abr{bert/electra}-large for
  reranking and span extraction. We include the retrieval results
  (though it gives slight gains) but do not compare the reader.  We
  expect the reader results are close to \beamdr{}, as both use
  similar models.}
%
%\halcomment{is beamdr really a baseline or is it more like a skyline? ditto mdr?}
%%cz0516 edited
%
%Both methods use passage labels as the annotation
%and claim better results over previous more complex multi-hop \abr{qa} systems. 
%

On \nq{}, we compare \name{} with two state-of-the-art dense retrieval
methods, \abr{dpr}~\cite{karpukhin-etal-2020-dense} and
\abr{ance}~\cite{xiong2020approximate}, which use same model architecture as
\abr{dpr}, with asynchronous negative evidence updates during
training.  We evaluate on fully-supervised and weakly-supervised
settings.
%, where the positives are Top BM-25 passages with answer existance.  \halcomment{do these need cites?} \halcomment{i don't really follow this previous sentence}
We directly use the released model checkpoint on full supervision and
train models from published code and data\footnote{The released data
  uses answers to match top \abr{bm}-25 results for distant supervision.}
on distant supervision.  All approaches use \bert{}-base as
pre-trained model.

%\czcomment{i feel the hotpotQA setting is still not very clear}
\begin{table}[!t]
  \centering
  \small
    \begin{tabular}{lccc}
      \toprule
       \textbf{Dataset}  & \textbf{Train}& \textbf{Dev}& \textbf{Test}\\
     \midrule
     \hotpot{} & 72,424 &5,918 & -  \\
      \nq{} & 79,168 &8,757 &3,610 \\
    \bottomrule
    \end{tabular}
    \caption{Number of questions on \hotpot{} and \nq{}. We use the dev (sub)set to evaluate \hotpot{}, 
    since the test set is hidden and we only use its bridge questions.
    }
  \label{tb:statistics}

\end{table}

\subsection{Implementation details}
On \hotpot{},
%We use \abr{bert}-base for both retriever and reader.  
%
we initialize our retriever with a 
dense retrieval checkpoint from \nq{} for both hops.\footnote{We make this choice because 
we need a cold starting point for our EM process (i.e., at the first iteration,
find some training signals from the retrieved evidence).
The initialization on \hotpot{} does not use any evidence labels 
(which would be cheating in our setting).
There are potentially alternative approaches to initializations that could be considered for future work.
}  
To initialize the reader, we first run \name{}'s retrieval for one iteration 
(without the reader filter), and train the reader from scratch, using the top-50 retrieval outputs.\footnote{We follow the reader setup from~\citet{karpukhin-etal-2020-dense}:
if multiple positives are in top-50 outputs, one of them is sampled as the positive instance in each training iteration.
Our pilot experiment shows that reader model is
 robust: such initialization gives reasonable accuracy.}
We use the same hyper-parameters as \beamdr{}, and train \name{} for eight 
iterations.\footnote{\name{} converges after five iterations (Figure~\ref{fig:analysis}).} 
%
%We use the same implementation as \name{} on both 
%reranking and span extraction, under distant supervision. 
%
%\halcomment{i think there's a subtle amount of "cheating" going on here: you limit yourself to hotpotqa questions with a two hops, right? and then you throw out that hop. and distdr specifically assumes two hop? in which case you're effectively using a small amount of supervision that you KNOW that the right number of hops.}
%\czcomment{Yes, that's correct. A even more general setting is to give a question, but does not give information of how many hops it requires. 
%A recent work from stanford tried to address this problem, but they focus on fully supervised setting by combining squad and hotpotQA for training. (so that setting is not perfect as well))}
%
%\halcomment{is using a checkpoint from nq cheating? if not, you should explain why not.}
%
%\czcomment{It's not, as init under distant sup setting, will fix the sentence.}
%
On \nq{}, we initialize \name{} from a \abr{dpr} checkpoint under distant supervision 
(so the evidence label is not used).
We train \name{} for ten iterations, using the same hyper-parameters as \abr{dpr}~\cite{karpukhin-etal-2020-dense}. 
We run \name{} on eight 2080Ti GPUs, and training takes three days.
%\todo{add significance testing}

\begin{table}[!tb]
  \small
  %\jbgcomment{Don't put space tweaks for camera ready, you'll get
  %  busted.  The table captions need to make it clearer what the
  %  numbers are and which models are being compared against.  E.g.,
  % ``Comparing DistDR against unsupervised models on retrieval.''
  % Make sure someone reading only table captions can get the picture.}
  %\vspace{-0.3cm}
  \footnotesize
  \begin{tabular}{m{3.5cm}	m{0.8cm} m{0.8cm} m{0.8cm}  }\toprule 
   Models  & Ans  & Passage & Chain  \\
    \midrule
    \multicolumn{4}{l}{\emph{No Supervision} }\\
    \abr{tf-idf} & 60.7 & 90.8 & 35.8\\
    \midrule
    \multicolumn{4}{l}{\emph{Full Supervision} }\\
    %\abr{grr}  & 87.8 & 93.3 & 77.9\\  %& 60.8\\ 
    \abr{mdr}  & 85.2 & 89.3 & \cellcolor{gray!50} 75.3\\  %& 60.8\\ 
    \beamdr{}  & 84.9 & 91.1 & 73.6\\% & 54.9\\
    \midrule
    \multicolumn{4}{l}{\emph{Distant Supervision} }\\
    \name{} w/ \abr{tf-idf} chains & 55.6 & 52.3& 22.3\\% &6.3\\
    \name{} & \cellcolor{gray!50} 86.2 &\cellcolor{gray!50} 92.2 & 75.1 \\%& 53.3\\ 
    %\multicolumn{5}{l}{\emph{Reranking from Retrieval Outputs}}\\
    %  \beamdr{}  & 90.5 & 94.7 & 83 & 68.9\\ 
    %  \name{} & 91.4 & 95.3 & 81.7 & 68.3\\ 
   
    \bottomrule
  \end{tabular}
      \caption{Compare \name{}'s retrieval (based on top-10 chains) with unsupervised and fully-supervised methods on \hotpot{} dev set
      over answer, passage and chain recall.
       \name{} matches fully-supervised dense retrieval approaches.}
      %Top: Retrieval from the whole corpus}
     % bottom: Reranking from top 100 full retrieval outputs. \cx{need more baseline. Separate supervised ones from distantly supervised ones.} \cx{I don't like the abbreviations of metric name, so hard to remember which is which.}}
      
      \label{tb:retrieval}
  
  \end{table}

\begin{table}[!tb]
    \small
      %\vspace{-0.3cm}
      \footnotesize
      \centering
      \begin{tabular}{m{2.5cm}	m{0.6cm} m{0.8cm} m{0.6cm} m{0.8cm}}\toprule 
       Models  &Ans &Passage& Chain& Span  \\
        \midrule
        \multicolumn{4}{l}{\emph{Full Supervision} }\\
        Transformer-XH & \cellcolor{gray!50} 91.8& \cellcolor{gray!50} 97.0 & 81.3 & 52.4 \\
        \grr{} & 87.5 & 92.2 & 79.0& 50.4\\
        \beamdr{} & 90.5 & 94.7 & \cellcolor{gray!50}83.0  & \cellcolor{gray!50}52.6 \\
        \midrule
        \multicolumn{4}{l}{\emph{Distant Supervision} }\\
        \name{} & 91.4 & 95.3 & 81.7 &51.6 \\ 
        \bottomrule
      \end{tabular}
          \caption{Compare \name{} with other fully-supervised methods with reranking over answer, passage and chain recall, and span extraction over span exact match. 
          \name{} is competitive to \beamdr{}. }
          
          \label{tb:ans_results}
      
      \end{table}

\begin{table}[!tb]
      %\vspace{-0.3cm}
      \small
      \begin{tabular}{m{3cm}	ccc  }\toprule 
       Models  & Top 1 & Top 20 & Span  \\
        \midrule
        \multicolumn{4}{l}{\emph{Full Supervision} }\\
        \dpr  & 46.3 & 78.4 & 41.5\\
        \abr{ance}    &\cellcolor{gray!50} 50.9 & \cellcolor{gray!50}81.9 &\cellcolor{gray!50} 46.0\\
        \midrule
        \multicolumn{4}{l}{\emph{Distant Supervision} }\\
        \dpr  & 45.2 & 78.2 &  37.9\\
        \abr{ance}    & 45.7 & 79.1 & 38.3\\
        \name{} & 50.4 & 80.1 & 40.5 \\\hline
      \end{tabular}
          \caption{Compare \name{} with other dense retrieval systems on \nq{} dataset over answer recall @1, 20, and span exact match.
          \name{} is competitive to fully-supervised approaches.}

          \label{tb:nq_retrieval}
      
      \end{table}

\subsection{Main Results}

Table~\ref{tb:retrieval} presents retrieval results on \hotpot{}. 
Here, \abr{tf-idf} is only able to find one evidence piece 
(usually the first hop which overlaps with question), but fails to
find all the evidence pieces.  Hence, using evidence from \abr{tf-idf} as distant supervision cannot
effectively train \name{}.
\name{} is slightly better than \beamdr{} and \abr{mdr}---both are state-of-the-art 
dense retrieval 
methods with full supervision---even though \name{} is only trained on (question, answer) pairs. 
%Through iterative updates, 
%\name{} increasingly retrieves reliable grounding evidence.
%that helps improve the model. %Following previous work, we also experiment on downstream reranking 
%and span extraction component. As our main scientific interest is not these absolute numbers per se, we 
%use the same model setting as \beamdr{}, but with distant supervision. 
When using
the same model implementation but with distant supervision, \name{} is
competitive to \beamdr{} on reader results (Table~\ref{tb:ans_results}).
%
%
%\todo{explain why bigger gap on reader (use annotated passage, which is different from 100 token splits)}
%
%\halcomment{ir or tf-idf?}
%cz0517 baseline uses BM25, so i would general say ir. 
On \nq{} (Table~\ref{tb:nq_retrieval}), unlike
multi-hop questions, using \abr{ir} to find evidence as distant supervision  provides helpful
training signals, which is confirmed by small gap between distant and
full supervision.\footnote{Compared to retrieval, the gap is larger for the reader.  This is due to 
training data processing for \abr{dpr} (and \abr{ance}).
%Since the original passages have been split and processed differently than 100-token split of candidate passages, 
On retrieval, \abr{dpr} replaces annotated gold passage with the corresponding 100-token passage in the candidate pool, 
and discard the questions if the matching is failed (25\% of questions). 
On reader, \abr{dpr} uses annotated passage as positive, and top retrieved 
passages that do not contain the answer are the negatives, thus entire training data is used. }
%On distant supervision, we use 74.2\% of training data for retrieval, but only 80.3\% training data is used for reader.}
Building on top of distantly-supervised \abr{dpr}, \name{} 
beats weakly-supervised systems, 
and is competitive with fully-supervised models. 
%Therefore, 
%\name{} is generally applicable to both single-hop \abr{qa} and multi-hop \abr{qa}, with on-par results to fully supervised \abr{sota} approaches.

%\halcomment{figure 3 is really too small; cannot read any of the stuff}
%cz0517 edited

\section{Analysis}
\label{sec:analysis}

This section explores \hotpot{} results to understand why \name{} is
on par with fully-supervised approaches.

\begin{table}[!tb]
    \small
      %\vspace{-0.3cm}
      \begin{tabular}{m{3.5cm}	m{0.6cm} m{0.7cm} m{0.7cm}  }\toprule 
       Models  & Ans & Passage & Chain \\
        \midrule
        \name{} & \cellcolor{gray!50} 86.2 &\cellcolor{gray!50} 92.2 &\cellcolor{gray!50} 75.1 \\ 
        \name{} w/ Sampled Pos & 83.9 & 90.9 & 72.1 \\ 
        Remove Reader Filter & 85.4 & 91.9  & 72.1\\ 
        Remove Non-gold Evidence & 66.7  & 74.1 & 53.1 \\ 
        \bottomrule
      \end{tabular}
          \caption{Ablation Studies evaluated on answer, passage and chain recall over top-10 chains. }
          
          \label{tb:ablation}
      
      \end{table}

\begin{figure*}[t]
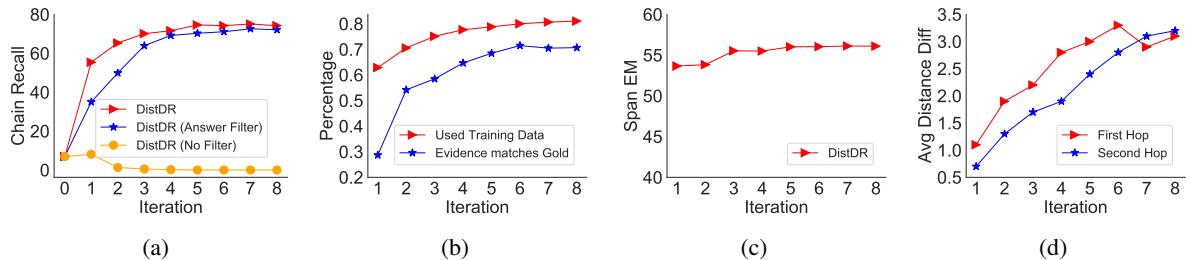

    \centering
     \begin{subfigure}{0.24\textwidth}
            \includegraphics[width=\textwidth]{\figfile{distdr-train.pdf}}
            \centering
            \caption{}
            \label{fig:analysis1}
        \end{subfigure}
        %\begin{subfigure}{0.24\textwidth}
        %    \includegraphics[width=\textwidth]{\figfile{trainrc.pdf}}
        %    \centering
        %    \caption{}
        %    \label{fig:analysis4}
        %\end{subfigure}
       \begin{subfigure}{0.24\textwidth}
            \includegraphics[width=\textwidth]{\figfile{trainex-comb.pdf}}
            \centering
            \caption{}
            \label{fig:analysis2}
        \end{subfigure}
        \begin{subfigure}{0.24\textwidth}
            \includegraphics[width=\textwidth]{\figfile{trainrc.pdf}}
            \centering
            \caption{}
            \label{fig:analysis4}
        \end{subfigure}
        \begin{subfigure}{0.24\textwidth}
            \includegraphics[width=\textwidth]{\figfile{dot-product.pdf}}
            \centering
            \caption{}
            %\caption{Distance Difference in dense space, \name{} increasingly larges the gap betweem Question postive and top 10 negative distances in dense space}
            \label{fig:analysis5}
        \end{subfigure}

    \caption{Quantitive analysis on \name{} by iteration. 
    (a): Compare different evidence filter strategies on dev set;  
    (b): Statistics on training set extracted evidence; (c): Compare span extraction component over gold evidence; 
    (d): Average distance difference on dev set from question to top-10 
    negative passages (Average) and positive passage. \name{} finds better evidence over iterations, the improved evidence further
    helps model training.
    \label{fig:analysis}
    %\jbgcomment{Add takeaway}
    }
    \end{figure*}

    %\begin{figure*}[t]
    %    \centering
    %    \small
    %     \begin{subfigure}{0.24\textwidth}
    %            \includegraphics[width=\textwidth]{\figfile{first_foo11.pdf}}
    %            \centering
    %            \caption{First Iteration}
    %            \label{fig:iter1}
    %        \end{subfigure}
    %       \begin{subfigure}{0.24\textwidth}
    %            \includegraphics[width=\textwidth]{\figfile{first_foo21.pdf}}
    %            \centering
    %            \caption{Second Iteration}
    %        \end{subfigure}
    %        \begin{subfigure}{0.24\textwidth}
    %            \includegraphics[width=\textwidth]{\figfile{first_foo31.pdf}}
    %            \centering
    %            \caption{Third Iteration}
    %            \label{fig:iter3}
    %        \end{subfigure}
    %        \begin{subfigure}{0.24\textwidth}
    %            \includegraphics[width=\textwidth]{\figfile{first_foo41.pdf}}
    %            \centering
    %            \caption{Fourth Iteration}
    %            \label{fig:iter4}
    %        \end{subfigure}
    %    \caption{Analysis on the vector distance of question and passage over iterations. 
    %    (a) - (d): First hop \abr{t-sne} visualization on one dev example with query~($Q$) and top-10 scored passage~($P$) embeddings on 
    %    first four iterations. 
    %    \label{fig:tsne}
    %  }
      %\jbgcomment{This figure isn't working for me yet.  I'ts not
      %  clear what a ``case visualization'' is.  Could you label one
      %  of the dots or something to make it less abstract?  Moreover,
      %  there needs to be a takeaway.}
   %     \end{figure*}

    \begin{figure*}[!t]
        \centering
        \small
        \includegraphics[width=0.99\linewidth]{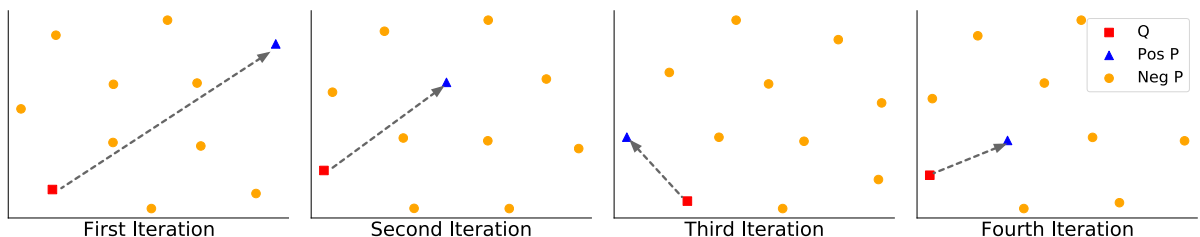}
        \caption{
         \abr{t-sne} visualization on an example with query~($Q$) and top-10 scored (first hop) passage~($P$) embeddings on 
        first four iterations. \name{} closes the distance between query and positive passage
        with more accurate evidence over iterations as distant supervision.}
        \label{fig:tsne}
        \end{figure*}

\subsection{Analysis on Model Training}
We first study how \name{} finds better evidence from a large corpus
with hard \abr{em}. 

%In this section, we conduct detailed analyses and ablation studies on \hotpot{}, to better understand how \name{} 
%iteratively learns to find evidence from a large corpus, and uses them as distant supervision for training. 
\begin{table*}[tb]
    \centering
  
    %\small
    \footnotesize

   \begin{tabular}{p{15cm}}
         \hline 
         \rowcolor{gray!50}
        \textbf{Q}: Jo Ann Terry won the 80m hurdles event at what Sao Paulo-based event from 1963?\\
        \rowcolor{Mycolor3}
        
        \textbf{A}: Pan American Games\\
        \rowcolor{Mycolor2}
        \textbf{Human Annotation}: Jo Ann Terry $\rightarrow$ 1963 pan american games
        
        \textbf{\name{}}: Jo Ann Terry $\rightarrow$ Jovem Pan \\

        \textbf{Jo Ann Terry}: Jo Ann Terry-Grissom  is a retired female hurdler from the United States. 
        Affiliated with the Tennessee State University she won the 80 m hurdles event at the 1963 Pan American Games.

        \textbf{1963 Pan American Games}: The 4th Pan American Games were held from April 20 to May 5, 1963, in São Paulo.

        \textbf{Jovem Pan}: Jovem Pan is the main Brazilian radio station based in São Paulo, Brazil.
        \\
        \rowcolor{gray!50}

        \textbf{Q}: Who's the Hungarian-born US film director renowned for adapting Stephen King novellas to the screen, including The Mist and The Green Mile?\\
        \rowcolor{Mycolor3}
        \textbf{A}: Frank Darabont \\
        \rowcolor{Mycolor2}
        \textbf{Human Annotation}: The Mist (film) $\rightarrow$ Frank Darabont
        
        \textbf{\name{}}: The Green Mile (film) $\rightarrow$ Frank Darabont\\

        \textbf{The Mist (film)}: The Mist is a 2007 American science-fiction horror film based on the 1980 novella "The Mist" by Stephen King. The film was written and directed by Frank Darabont. 

        \textbf{The Green Mile (film)}: The Green Mile is a 1999 American fantasy crime drama film written and directed by Frank Darabont and adapted from the 1996 Stephen King novel of the same name.

        \textbf{Frank Darabont}: Frank Arpad Darabont is a French-Hungarian-American film director, screenwriter and producer.
         As a director he is known for his film adaptations of Stephen King novellas such as "The Green Mile" and "The Mist".
        \\
        \rowcolor{gray!50}
          \textbf{Q}: Zimbabwe's Guwe Secondary School has a sister school in what New York county?\\
          \rowcolor{Mycolor3}
        
        \textbf{A}: Nassau County \\
        \rowcolor{Mycolor2}
        \textbf{Human Annotation}: Guwe Secondary School $\rightarrow$ Carle Place High School
        
        \textbf{\name{}}: University of Zimbabwe $\rightarrow$ East Rockaway High School\\

        \textbf{Guwe Secondary School}: Guwe Secondary School is located in Zimbabwe. It has a sister school 
        in Carle Place, New York, United States.

        \textbf{Carle Place High School}: Carle Place Middle/High School is a six-year comprehensive public high school 
        located in the hamlet of Carle Place in Nassau County, New York.

        \textbf{University of Zimbabwe}: The University of Zimbabwe (UZ) is the oldest and formerly largest university in Zimbabwe.

        \textbf{East Rockaway High School}: East Rockaway Junior-Senior High School is a co-educational six-year secondary school in East Rockaway, 
        New York, and the sole high school in Nassau County, New York.
        \\\hline 
        \caption{Case study of \name{} on ``false positive'' evidence
     from \hotpot{}
     that does not match gold evidence. Despite the dataset goals,
     some examples are answerable with a single piece of evidence
     evidence (top).  Other times, \name{} finds alternate valid
     evidence (middle).  However, it can also find incorrect evidence (bottom).}
    \label{tb:case}
  
    \end{tabular}
  \end{table*}

\paragraph{Effect of Hard EM} 
We study the effect of hard \abr{em} by going through accuracy and 
extracted evidence statistics at every iteration. 
Chain recall for \name{} on the dev set increases
over the first five iterations (Figure~\ref{fig:analysis1}) and then
converges. As the evidence finder improves, the E-step extracts better
evidence for training examples (Figure~\ref{fig:analysis2}):  
both the percentage of used training examples and gold evidence in training examples 
increase with additional \abr{em} iterations.  The improved evidence quality further 
helps \name{} training. We compare multiple positive sampling strategies:
using hard \abr{em} (top-1 as positive evidence) is slightly better than randomly
sampling from top-$k$ positives for the M-step update (Table~\ref{tb:ablation}, second row). 
%\halcomment{this prev sentence is too long; please split up to make more clear}
%% cz0517 edited

%Maximum Marginal Likelikehood strategy. 

\paragraph{The Effect of Evidence Filtering}
Filtering positive evidence is crucial for ensuring high-quality
evidence. We ablate different filtering methods in Table~\ref{tb:ablation} 
and Figure~\ref{fig:analysis1}: No filter, only an answer filter, and
an additional reader filter (\name{}).
%
%Table~\ref{tb:ablation} and Figure~\ref{fig:analysis1} compare
%different methods:
%
No filter fails completely: evidence should not be considered correct
if it does not contain the answer.
With an additional reader filter, \name{} outperforms using only the
answer matching filter and converges faster (since the reader filter
reduces false-positives thus makes training more robust).

Despite rapidly changing evidence, the reader's span extraction component is 
robust over iterations (Figure~\ref{fig:analysis4}), even on the first iteration, suggesting it is
robust against spurious false-positive evidence.

%\tablefile{human-eval}

%\halcomment{text in figure 4 are all too small to read. also the caption should tell me the take-aways from both figures}
%cz0517 edited
\paragraph{Visualizing \name{} in Dense Space}
%In this section, we conduct quantative and qualitative analysis to understand \name{}'s
%behavior in dense space. 
\name{} improves evidence retrieval by bringing the
question representation closer to positive evidence and pushing
negative evidence away (Figure~\ref{fig:analysis5}).  
In Figure~\ref{fig:tsne}, \abr{t-sne}~\cite{maaten2008visualizing} visualizes 
the first hop query and passage representations 
(second hop representations are similar) in dense space at each
iteration. As expected, 
at the beginning, the question is far from the positive passage, 
with negative passages between. After updating the representations,
the distance gets closer until positive passage is the closest to the question.

%\subsection{Case Study}
\subsection{Analysis on Retrieved Evidence}
\label{subsec:evi}

\name{} is competitive to fully-supervised approaches on \hotpot{}.
However, even at the last iteration, only $65\%$ of extracted 
evidence matches the gold.
We ablate evidence and contrasts gold evidence with that
retrieved by \name{}.

\paragraph{Non-gold Evidence is Helpful}
If we only retain training examples that 
matches the labeled gold evidence, answer accuracy falters and 
underperform \name{} (Table~\ref{tb:ablation}, last row).
Instead of providing noise, non-gold evidence gives
useful signal for model training.

%\halcomment{i know you're putting false positive in quotes, but i would either explicitly explain the quotes, or pick a different term, like "alternative evidence" or something.}
% cz0517 changed to non-gold

\paragraph{Human Analysis and Case Study}
We manually annotate fifty training examples where the extracted 
evidence from \name{} does not match the labeled evidence.
We confirm that 
most of the extracted evidence is helpful for model training.
Specifically, 
$38\%$ of the questions are answerable via a single piece of evidence, even
though this dataset is supposed to require multiple hops
(Table~\ref{tb:case}).
In another $28\%$ of cases, \name{} finds alternate valid evidence.
In the second example in Table~\ref{tb:case}, the question mentions both films \underline{The Mist}
and \underline{The Green Mile}; therefore the reasoning chain from either film to the director
\underline{Frank Darabont} is correct (though only one is annotated).
In the final of $34\%$ of cases, \name{} finds the wrong evidence, often because the extracted evidence only includes one span that 
matches the answer type, therefore the reader confidently outputs the
span for the wrong reason.
In the third example in Table~\ref{tb:case}, \ul{Nassau County} is the
only county it sees, and therefore the model has stumbled upon the right
answer erroneously.
Building a model with faithful predictions is an important 
ongoing research topic~\cite{jacovi-goldberg-2020-towards}.

%\halcomment{i think i would cut table 5, all of the information is in the text and better explained there.}

\section{Related Work}
\label{sec:related}

\paragraph{Question Answering Datasets}
There is growing interest in \abr{nlp} communities to build 
large-scale datasets~\cite[\emph{inter
alia}]{rajpurkar-etal-2016-squad, jia2017adversarial, dua2019drop} 
for \abr{qa} research. In addition to questions and answers, benchmark datasets 
often include annotated evidence, but it requires significant 
 annotation protocol design and human annotations.  
In \abr{squad}~\cite{rajpurkar-etal-2016-squad}, among the first large-scale
reading comprehension datasets, annotators write questions conditioned
on a passage, which creates dataset artifacts~\cite{jia2017adversarial}. 
To overcome dataset artifacts, \nq{}~\cite{kwiatkowski2019natural} 
%treat \abr{qa} as an information seeking task, 
use real Google queries as questions and ask annotators to 
label both evidence passages and short answers. But such annotation is expensive,
as they further ask additional experts to verify whether the evidence correctly leads to the answer.
Instead, \name{} focuses on the weakly-supervised setting with only
question--answer pairs, which is significantly cheaper.

Annotation is more fraught for multi-hop \abr{qa} datasets~\cite{yang+18b}.
To construct \hotpot{}, annotators are
presented with a linked Wikipedia passages, as 
pilot studies indicate that it is difficult to ask a meaningful multi-hop question with arbitrary passages. 
However, some questions in \hotpot{} 
include shortcuts that are answerable by a single passage~\cite{min2019compositional}, which is confirmed by our analysis (Section~\ref{subsec:evi}).

Distant supervision has been successfully adopted for many \abr{nlp} tasks such as 
relation extraction~\cite{mintz-etal-2009-distant}.
Recent work builds \abr{qa} datasets with distant supervision, such as
\abr{TriviaQA}~\cite{joshi-etal-2017-triviaqa},
\abr{SearchQA}~\cite{dunn2017searchqa},
\abr{QBLink}~\cite{Elgohary-18} by automatically gathering evidence
documents from a corpus as distant supervision for available
question-answer pairs.
They use standard \abr{ir} techniques to find relevant passages and match them with answer strings. 
These methods succeed for simple questions where terms overlap with evidence passages, 
%(though falters on evidence that requires ``semantic matching''~\citep{karpukhin-etal-2020-dense}). 
This no longer holds for multi-hop questions that require a 
reasoning chain as evidence to the answer: evidence pieces do not
overlap with the question but rather depend on the previous evidence
pieces.
Unsupervised \abr{ir} methods cannot capture such implicit relations. 
\name{} removes the burden of little textual overlap through dense retrieval and its
iterative process retrieves better evidence. 

%\jbgcomment{These cite dumps aren't helpful.  Focus on stuff not cited
%elsewhere and be more specific}

\paragraph{Open-domain \abr{qa} systems}
\citet{chen2017reading} first combine information retrieval and (neural) reading comprehension 
for open-domain \abr{qa}. Several works aim to improve the neural reader~\cite[\emph{inter
alia}]{clark-gardner-2018-simple, wang2017rtheta3}, or use generative
models to compose an answer~\cite[\emph{inter
alia}]{lewis2020retrieval, izacard-grave-2021-leveraging}. Recent progress~\cite[\emph{inter
alia}]{karpukhin-etal-2020-dense,xiong2020approximate,xiong2020answering} uses dense retrieval 
to aid both single-hop 
and multi-hop questions. 
However, a crucial distinction 
is that these approaches assume the evidence 
%(either from annotators or automatically retrieved) 
is given for training, while \name{} iteratively finds evidence 
and uses it to improve the model.
\citet{min2019discrete} also use hard-\abr{em} for weakly-supervised \abr{qa}, but---orthogonal 
to our approach---they assume the evidence is given and
find the most likely answer mentions in the evidence, 
while we aim to find evidence from a large corpus. 

%\paragraph{Distant Supervision in NLP}

\section{Conclusion}
\label{sec:conclu}

We present \name{}, a distantly-supervised \abr{odqa} system that improves over a 
weak retriever by iteratively
finding evidence from a corpus, and using the 
evidence as distant supervision for model training. Without using any evidence labels, 
\name{} matches the fully-supervised \abr{sota} approaches on both multi-hop and single-hop \abr{qa}
benchmarks.

Annotating evidence for existing question-answer pairs is generally expensive, especially for complex questions.
While \name{} can accurately find evidence for arbitrary complex
machine reading-style questions, future work needs to validate whether
this can work for other types of questions.
This could improve the reader to answer numerical reasoning~\cite{dua2019drop}, 
%commonsense reasoning~\cite{talmor2019commonsenseqa}, 
temporal reasoning~\cite{ning2020torque}, multi-model reasoning~\cite{lei-18}, 
or combination of these skills~\cite{bartolo2020beat}.
%~\cite{Rodriguez:Feng:Iyyer:He:Boyd-Graber-Preprint}.

%\jbgcomment{Conclusion would be stronger with more future work.  E.g.,
%consider multihop questions written by experts (qb/qb-link), use the
%same ideas to detect ambiguity, etc.}

\section*{Acknowledgments}

We thank CLIP members, Tianze Shi, anonymous reviewers
and meta-reviewer
for their suggestions and comments.
Zhao is supported by the Office of the Director of National Intelligence (\abr{odni}), 
Intelligence Advanced Research Projects Activity (\abr{iarpa}), via the \abr{better} Program contract 2019-19051600005. 
Boyd-Graber is supported by \abr{nsf} Grant IIS-1822494.
Any opinions, findings, conclusions, or recommendations
expressed here are those of the authors and do not
necessarily reflect the view of the sponsors.

\bibliographystyle{style/acl_natbib}
\bibliography{bib/journal-full,bib/ref}

\begin{thebibliography}{34}
\expandafter\ifx\csname natexlab\endcsname\relax\def\natexlab#1{#1}\fi

\bibitem[{Asai et~al.(2020)Asai, Hashimoto, Hajishirzi, Socher, and
  Xiong}]{asai2020learning}
Akari Asai, Kazuma Hashimoto, Hannaneh Hajishirzi, Richard Socher, and Caiming
  Xiong. 2020.
\newblock \href {https://openreview.net/forum?id=SJgVHkrYDH} {Learning to
  retrieve reasoning paths over {W}ikipedia graph for question answering}.
\newblock In \emph{Proceedings of the International Conference on Learning
  Representations}.

\bibitem[{Bartolo et~al.(2020)Bartolo, Roberts, Welbl, Riedel, and
  Stenetorp}]{bartolo2020beat}
Max Bartolo, Alastair Roberts, Johannes Welbl, Sebastian Riedel, and Pontus
  Stenetorp. 2020.
\newblock \href
  {https://direct.mit.edu/tacl/article/doi/10.1162/tacl_a_00338/96474/Beat-the-AI-Investigating-Adversarial-Human}
  {Beat the ai: Investigating adversarial human annotation for reading
  comprehension}.
\newblock \emph{Transactions of the Association for Computational Linguistics}.

\bibitem[{Chen et~al.(2017)Chen, Fisch, Weston, and Bordes}]{chen2017reading}
Danqi Chen, Adam Fisch, Jason Weston, and Antoine Bordes. 2017.
\newblock \href {https://www.aclweb.org/anthology/P17-1171/} {Reading
  {W}ikipedia to answer open-domain questions}.
\newblock In \emph{Proceedings of the Association for Computational
  Linguistics}.

\bibitem[{Cheng et~al.(2020)Cheng, Chang, Lee, and
  Toutanova}]{cheng-etal-2020-probabilistic}
Hao Cheng, Ming-Wei Chang, Kenton Lee, and Kristina Toutanova. 2020.
\newblock \href {https://doi.org/10.18653/v1/2020.acl-main.501} {Probabilistic
  assumptions matter: Improved models for distantly-supervised document-level
  question answering}.
\newblock In \emph{Proceedings of the Association for Computational
  Linguistics}.

\bibitem[{Clark and Gardner(2018)}]{clark-gardner-2018-simple}
Christopher Clark and Matt Gardner. 2018.
\newblock \href {https://doi.org/10.18653/v1/P18-1078} {Simple and effective
  multi-paragraph reading comprehension}.
\newblock In \emph{Proceedings of the Association for Computational
  Linguistics}.

\bibitem[{Dua et~al.(2019)Dua, Wang, Dasigi, Stanovsky, Singh, and
  Gardner}]{dua2019drop}
Dheeru Dua, Yizhong Wang, Pradeep Dasigi, Gabriel Stanovsky, Sameer Singh, and
  Matt Gardner. 2019.
\newblock \href {https://www.aclweb.org/anthology/N19-1246} {Drop: A reading
  comprehension benchmark requiring discrete reasoning over paragraphs}.
\newblock In \emph{Conference of the North American Chapter of the Association
  for Computational Linguistics}.

\bibitem[{Dunn et~al.(2017)Dunn, Sagun, Higgins, Guney, Cirik, and
  Cho}]{dunn2017searchqa}
Matthew Dunn, Levent Sagun, Mike Higgins, V~Ugur Guney, Volkan Cirik, and
  Kyunghyun Cho. 2017.
\newblock \href {https://arxiv.org/abs/1704.05179} {Search{QA}: A new q\&a
  dataset augmented with context from a search engine}.
\newblock \emph{arXiv preprint arXiv:1704.05179}.

\bibitem[{Elgohary et~al.(2018)Elgohary, Zhao, and Boyd-Graber}]{Elgohary-18}
Ahmed Elgohary, Chen Zhao, and Jordan Boyd-Graber. 2018.
\newblock \href {https://www.aclweb.org/anthology/D18-1134/} {Dataset and
  baselines for sequential open-domain question answering}.
\newblock In \emph{Proceedings of Empirical Methods in Natural Language
  Processing}.

\bibitem[{Guu et~al.(2020)Guu, Lee, Tung, Pasupat, and Chang}]{guu2020realm}
Kelvin Guu, Kenton Lee, Zora Tung, Panupong Pasupat, and Ming-Wei Chang. 2020.
\newblock \href {https://arxiv.org/abs/2002.08909} {{REALM}:
  Retrieval-augmented language model pre-training}.
\newblock In \emph{Proceedings of the International Conference of Machine
  Learning}.

\bibitem[{Izacard and Grave(2021)}]{izacard-grave-2021-leveraging}
Gautier Izacard and Edouard Grave. 2021.
\newblock \href {https://aclanthology.org/2021.eacl-main.74} {Leveraging
  passage retrieval with generative models for open domain question answering}.
\newblock In \emph{Proceedings of the European Chapter of the Association for
  Computational Linguistics}.

\bibitem[{Jacovi and Goldberg(2020)}]{jacovi-goldberg-2020-towards}
Alon Jacovi and Yoav Goldberg. 2020.
\newblock \href {https://www.aclweb.org/anthology/2020.acl-main.386} {Towards
  faithfully interpretable {NLP} systems: How should we define and evaluate
  faithfulness?}
\newblock In \emph{Proceedings of the Association for Computational
  Linguistics}.

\bibitem[{Jia and Liang(2017)}]{jia2017adversarial}
Robin Jia and Percy Liang. 2017.
\newblock \href {https://www.aclweb.org/anthology/D17-1215/} {Adversarial
  examples for evaluating reading comprehension systems}.
\newblock In \emph{Proceedings of Empirical Methods in Natural Language
  Processing}.

\bibitem[{Joshi et~al.(2017)Joshi, Choi, Weld, and
  Zettlemoyer}]{joshi-etal-2017-triviaqa}
Mandar Joshi, Eunsol Choi, Daniel Weld, and Luke Zettlemoyer. 2017.
\newblock \href {https://doi.org/10.18653/v1/P17-1147} {{T}rivia{QA}: A large
  scale distantly supervised challenge dataset for reading comprehension}.
\newblock In \emph{Proceedings of the Association for Computational
  Linguistics}.

\bibitem[{Karpukhin et~al.(2020)Karpukhin, Oguz, Min, Lewis, Wu, Edunov, Chen,
  and Yih}]{karpukhin-etal-2020-dense}
Vladimir Karpukhin, Barlas Oguz, Sewon Min, Patrick Lewis, Ledell Wu, Sergey
  Edunov, Danqi Chen, and Wen-tau Yih. 2020.
\newblock \href {https://www.aclweb.org/anthology/2020.emnlp-main.550} {Dense
  passage retrieval for open-domain question answering}.
\newblock In \emph{Proceedings of Empirical Methods in Natural Language
  Processing}.

\bibitem[{Kwiatkowski et~al.(2019)Kwiatkowski, Palomaki, Redfield, Collins,
  Parikh, Alberti, Epstein, Polosukhin, Devlin, Lee
  et~al.}]{kwiatkowski2019natural}
Tom Kwiatkowski, Jennimaria Palomaki, Olivia Redfield, Michael Collins, Ankur
  Parikh, Chris Alberti, Danielle Epstein, Illia Polosukhin, Jacob Devlin,
  Kenton Lee, et~al. 2019.
\newblock \href
  {https://storage.googleapis.com/pub-tools-public-publication-data/pdf/1f7b46b5378d757553d3e92ead36bda2e4254244.pdf}
  {Natural questions: a benchmark for question answering research}.
\newblock \emph{Transactions of the Association for Computational Linguistics}.

\bibitem[{Lee et~al.(2019)Lee, Chang, and Toutanova}]{lee-etal-2019-latent}
Kenton Lee, Ming-Wei Chang, and Kristina Toutanova. 2019.
\newblock \href {https://www.aclweb.org/anthology/P19-1612} {Latent retrieval
  for weakly supervised open domain question answering}.
\newblock In \emph{Proceedings of the Association for Computational
  Linguistics}.

\bibitem[{Lei et~al.(2018)Lei, Yu, Bansal, and Berg}]{lei-18}
Jie Lei, Licheng Yu, Mohit Bansal, and Tamara Berg. 2018.
\newblock \href {https://www.aclweb.org/anthology/D18-1167/} {{TVQA}:
  Localized, compositional video question answering}.
\newblock In \emph{Proceedings of Empirical Methods in Natural Language
  Processing}.

\bibitem[{Lewis et~al.(2020)Lewis, Perez, Piktus, Petroni, Karpukhin, Goyal,
  K{\"u}ttler, Lewis, Yih, Rockt{\"a}schel et~al.}]{lewis2020retrieval}
Patrick Lewis, Ethan Perez, Aleksandra Piktus, Fabio Petroni, Vladimir
  Karpukhin, Naman Goyal, Heinrich K{\"u}ttler, Mike Lewis, Wen-tau Yih, Tim
  Rockt{\"a}schel, et~al. 2020.
\newblock \href
  {https://proceedings.neurips.cc/paper/2020/file/6b493230205f780e1bc26945df7481e5-Paper.pdf}
  {Retrieval-augmented generation for knowledge-intensive nlp tasks}.
\newblock In \emph{Proceedings of Advances in Neural Information Processing
  Systems}.

\bibitem[{Maaten and Hinton(2008)}]{maaten2008visualizing}
Laurens van~der Maaten and Geoffrey Hinton. 2008.
\newblock \href {https://www.jmlr.org/papers/v9/vandermaaten08a.html}
  {Visualizing data using t-{SNE}}.
\newblock \emph{Journal of machine learning research}, 9(11).

\bibitem[{Min et~al.(2019{\natexlab{a}})Min, Chen, Hajishirzi, and
  Zettlemoyer}]{min2019discrete}
Sewon Min, Danqi Chen, Hannaneh Hajishirzi, and Luke Zettlemoyer.
  2019{\natexlab{a}}.
\newblock \href {https://www.aclweb.org/anthology/D19-1284/} {A discrete hard
  {EM} approach for weakly supervised question answering}.
\newblock In \emph{Proceedings of Empirical Methods in Natural Language
  Processing}.

\bibitem[{Min et~al.(2019{\natexlab{b}})Min, Wallace, Singh, Gardner,
  Hajishirzi, and Zettlemoyer}]{min2019compositional}
Sewon Min, Eric Wallace, Sameer Singh, Matt Gardner, Hannaneh Hajishirzi, and
  Luke Zettlemoyer. 2019{\natexlab{b}}.
\newblock \href {https://www.aclweb.org/anthology/P19-1416/} {Compositional
  questions do not necessitate multi-hop reasoning}.
\newblock In \emph{Proceedings of the Association for Computational
  Linguistics}.

\bibitem[{Mintz et~al.(2009)Mintz, Bills, Snow, and
  Jurafsky}]{mintz-etal-2009-distant}
Mike Mintz, Steven Bills, Rion Snow, and Daniel Jurafsky. 2009.
\newblock \href {https://www.aclweb.org/anthology/P09-1113} {Distant
  supervision for relation extraction without labeled data}.
\newblock In \emph{Proceedings of the Association for Computational
  Linguistics}.

\bibitem[{Ning et~al.(2020)Ning, Wu, Han, Peng, Gardner, and
  Roth}]{ning2020torque}
Qiang Ning, Hao Wu, Rujun Han, Nanyun Peng, Matt Gardner, and Dan Roth. 2020.
\newblock \href {https://aclanthology.org/2020.emnlp-main.88} {Torque: A
  reading comprehension dataset of temporal ordering questions}.
\newblock In \emph{Proceedings of Empirical Methods in Natural Language
  Processing}.

\bibitem[{Rajpurkar et~al.(2016)Rajpurkar, Zhang, Lopyrev, and
  Liang}]{rajpurkar-etal-2016-squad}
Pranav Rajpurkar, Jian Zhang, Konstantin Lopyrev, and Percy Liang. 2016.
\newblock \href {https://doi.org/10.18653/v1/D16-1264} {{SQ}u{AD}: 100,000+
  questions for machine comprehension of text}.
\newblock In \emph{Proceedings of Empirical Methods in Natural Language
  Processing}.

\bibitem[{Russell(2019)}]{russell-19}
Daniel~M Russell. 2019.
\newblock \href {https://mitpress.mit.edu/books/joy-search} {\emph{The Joy of
  Search: A Google Insider's Guide to Going Beyond the Basics}}.
\newblock MIT Press.

\bibitem[{Samdani et~al.(2012)Samdani, Chang, and Roth}]{samdani2012unified}
Rajhans Samdani, Ming-Wei Chang, and Dan Roth. 2012.
\newblock \href {https://www.aclweb.org/anthology/N12-1087.pdf} {Unified
  expectation maximization}.
\newblock In \emph{Conference of the North American Chapter of the Association
  for Computational Linguistics}.

\bibitem[{Shrivastava and Li(2014)}]{shrivastava2014asymmetric}
Anshumali Shrivastava and Ping Li. 2014.
\newblock \href {https://dl.acm.org/doi/10.5555/2969033.2969086} {Asymmetric
  {LSH} ({ALSH}) for sublinear time maximum inner product search ({MIPS})}.
\newblock In \emph{Proceedings of Advances in Neural Information Processing
  Systems}.

\bibitem[{Voorhees et~al.(1999)}]{voorhees1999trec}
Ellen~M Voorhees et~al. 1999.
\newblock \href
  {https://www.nist.gov/publications/trec-8-question-answering-track-report}
  {The {TREC}-8 question answering track report}.
\newblock In \emph{Proceedings of Text Retrieval Conference}.

\bibitem[{Wang et~al.(2018)Wang, Yu, Guo, Wang, Klinger, Zhang, Chang, Tesauro,
  Zhou, and Jiang}]{wang2017rtheta3}
Shuohang Wang, Mo~Yu, Xiaoxiao Guo, Zhiguo Wang, Tim Klinger, Wei Zhang, Shiyu
  Chang, Gerald Tesauro, Bowen Zhou, and Jing Jiang. 2018.
\newblock \href {https://arxiv.org/abs/1709.00023} {R$^3$: Reinforced
  reader-ranker for open-domain question answering}.
\newblock In \emph{Proceedings of the Association for the Advancement of
  Artificial Intelligence}.

\bibitem[{Xiong et~al.(2021{\natexlab{a}})Xiong, Xiong, Li, Tang, Liu, Bennett,
  Ahmed, and Overwijk}]{xiong2020approximate}
Lee Xiong, Chenyan Xiong, Ye~Li, Kwok-Fung Tang, Jialin Liu, Paul Bennett,
  Junaid Ahmed, and Arnold Overwijk. 2021{\natexlab{a}}.
\newblock \href {https://openreview.net/forum?id=zeFrfgyZln} {Approximate
  nearest neighbor negative contrastive learning for dense text retrieval}.
\newblock In \emph{Proceedings of the International Conference on Learning
  Representations}.

\bibitem[{Xiong et~al.(2021{\natexlab{b}})Xiong, Li, Iyer, Du, Lewis, Wang,
  Mehdad, tau Yih, Riedel, Kiela, and Oğuz}]{xiong2020answering}
Wenhan Xiong, Xiang~Lorraine Li, Srini Iyer, Jingfei Du, Patrick Lewis,
  William~Yang Wang, Yashar Mehdad, Wen tau Yih, Sebastian Riedel, Douwe Kiela,
  and Barlas Oğuz. 2021{\natexlab{b}}.
\newblock \href {https://openreview.net/forum?id=EMHoBG0avc1} {Answering
  complex open-domain questions with multi-hop dense retrieval}.
\newblock In \emph{Proceedings of the International Conference on Learning
  Representations}.

\bibitem[{Yang et~al.(2018)Yang, Qi, Zhang, Bengio, Cohen, Salakhutdinov, and
  Manning}]{yang+18b}
Zhilin Yang, Peng Qi, Saizheng Zhang, Yoshua Bengio, William Cohen, Ruslan
  Salakhutdinov, and Christopher~D. Manning. 2018.
\newblock \href {https://doi.org/10.18653/v1/D18-1259} {{{HotpotQA}}: A dataset
  for diverse, explainable multi-hop question answering}.
\newblock In \emph{Proceedings of Empirical Methods in Natural Language
  Processing}.

\bibitem[{Zhao et~al.(2021)Zhao, Xiong, Boyd-Graber, and
  Daum{\'e}~III}]{zhao2021multi}
Chen Zhao, Chenyan Xiong, Jordan Boyd-Graber, and Hal Daum{\'e}~III. 2021.
\newblock \href {https://arxiv.org/abs/2104.05883} {Multi-step reasoning over
  unstructured text with beam dense retrieval}.
\newblock In \emph{Conference of the North American Chapter of the Association
  for Computational Linguistics}.

\bibitem[{Zhao et~al.(2020)Zhao, Xiong, Rosset, Song, Bennett, and
  Tiwary}]{zhaotransxh2020}
Chen Zhao, Chenyan Xiong, Corby Rosset, Xia Song, Paul Bennett, and Saurabh
  Tiwary. 2020.
\newblock \href {https://openreview.net/forum?id=r1eIiCNYwS} {Transformer-xh:
  Multi-evidence reasoning with extra hop attention}.
\newblock In \emph{Proceedings of the International Conference on Learning
  Representations}.

\end{thebibliography}

\end{document}